\documentclass{article}
\usepackage{iclr2027_conference,times}
\usepackage[T1]{fontenc}
\usepackage[utf8]{inputenc}
\usepackage{amsmath,amssymb,mathtools,bm}
\usepackage{graphicx,booktabs,array,tabularx,multirow}
\usepackage{xcolor}
\usepackage{microtype}
\usepackage{url}
\usepackage[hidelinks]{hyperref}
\usepackage{placeins,flafter}
\usepackage{catchfile}
\iclrfinalcopy
\title{WindINR: Latent-State INR for Fast\\
Local Wind Query and Correction\\
in Complex Terrain}
\author{%
Yi Xiao\textsuperscript{1,5}, Qilong Jia\textsuperscript{1}, Hang Fan\textsuperscript{2}, Pascal Fua\textsuperscript{3},\\
\bfseries Robert Jenssen\textsuperscript{4}, Xiaosong Ma\textsuperscript{5}, Wei Xue\textsuperscript{1}\\[0.6em]
\normalfont\small\textsuperscript{1}\,Tsinghua University\\
\normalfont\small\textsuperscript{2}\,Columbia University\\
\normalfont\small\textsuperscript{3}\,EPFL\\
\normalfont\small\textsuperscript{4}\,The Arctic University of Norway\\
\normalfont\small\textsuperscript{5}\,Mohamed bin Zayed University of Artificial Intelligence
}
\newcommand{\R}{\mathbb{R}}

\newcommand{\norm}[1]{\left\lVert#1\right\rVert}
\newcommand{\ms}{\mathrm{m\,s^{-1}}}

\newcommand{\tromso}{Troms\o{}}
\hypersetup{pdftitle={WindINR: Latent-State INR for Fast Local Wind Query and Correction in Complex Terrain},pdfauthor={Yi Xiao, Qilong Jia, Hang Fan, Pascal Fua, Robert Jenssen, Xiaosong Ma, Wei Xue},pdfsubject={arXiv preprint}}

\begin{document}
\maketitle
% The preprint makes no conference review or acceptance claim.
\lhead{}
\begin{abstract}
Many downstream decisions in complex terrain require fast wind estimates at a small number of user-specified locations and heights for a given forecast valid time, rather than another dense forecast field on a fixed grid. We present WindINR, a latent-state implicit neural representation framework for continuous high-resolution local wind query and sparse-observation correction. WindINR maps static terrain descriptors, a low-resolution background field, and continuous query coordinates to a high-resolution wind state through a latent-conditioned decoder. To enable rapid inference-time correction, WindINR separates reusable representation learning from sample-specific latent-state correction. During training, a privileged encoder infers a reference latent state from high-resolution supervision, a deployable latent predictor estimates an initial latent state from inference-time inputs alone, and their discrepancies are summarized into a dataset-adaptive Gaussian prior over latent corrections. At inference time, within the WindINR module, network weights remain fixed and only the latent state is updated by minimizing a regularized correction objective using sparse observations and their uncertainty. We train WindINR on the large-scale FuXi-CFD dataset spanning diverse terrain and evaluate it across complementary observation settings, including controlled OSSEs, simulated helicopter trajectories, and real UAV measurements over Tromsø. Across these settings, assimilating sparse observations consistently improves local wind-field predictions, with WindINR outperforming several baselines, including WindSeer. Moreover, by performing online correction only in a low-dimensional latent space, WindINR achieves a 14$\times$ speedup over full-network fine-tuning. The resulting representation remains continuously queryable at arbitrary spatial coordinates, making WindINR a promising and efficient correction framework that provides a practical interface between kilometer-scale background products, sparse local observations, and wind queries in complex terrain.
\end{abstract}

\section{Introduction}
\label{sec:introduction}
\begin{figure}[t]
\centering
\includegraphics[width=0.95\linewidth]{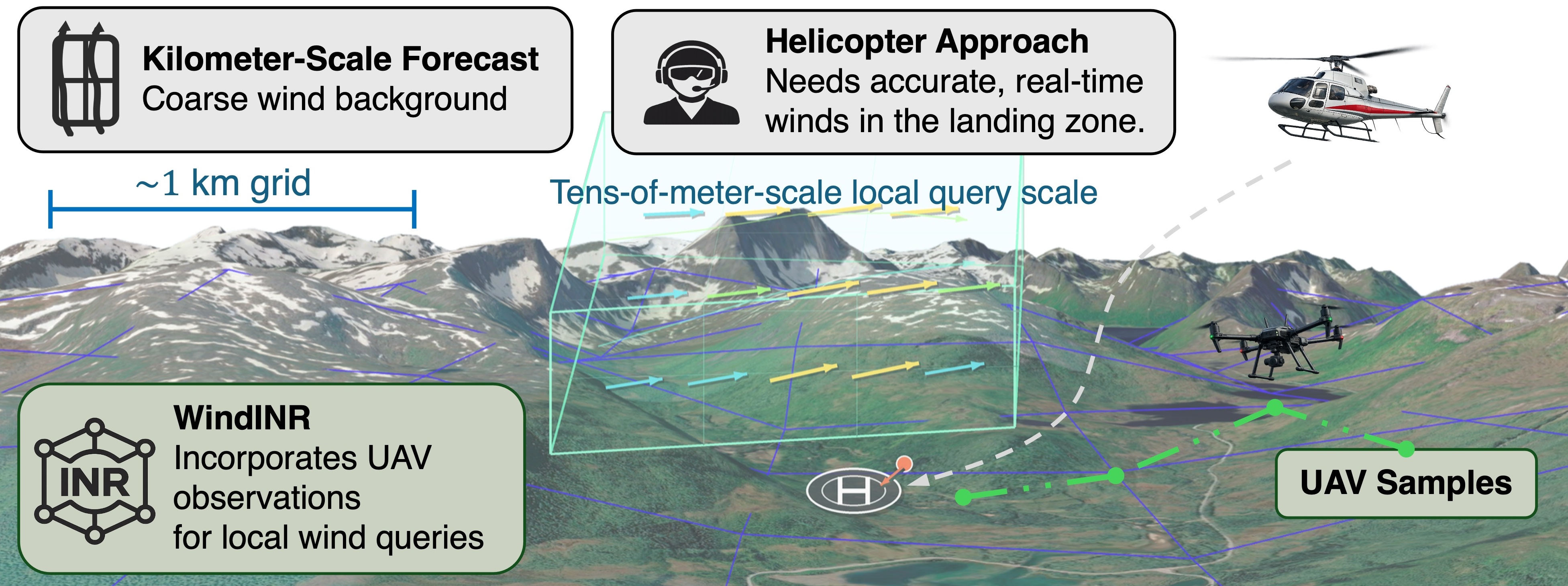}
\caption{\textbf{Operational motivation for WindINR.} Kilometer-scale forecast products provide useful coarse background information over complex terrain, but helicopter approach and landing decisions require fast wind estimates at tens-of-meters scale and at user-specified off-grid locations. In the illustrated workflow, a UAV flying ahead of the helicopter provides sparse local wind observations along the approach corridor. WindINR uses the terrain and coarse forecast background as conditioning inputs, and rapidly updates a compact latent state with these sparse observations while preserving continuous local wind query.}
\label{fig:motivation}
\end{figure}

Operational weather products are usually delivered as gridded analyses and forecasts, but many decisions in complex terrain require only a small number of fast, accurate wind estimates at user-specified locations. During a helicopter approach, for example, the relevant question is often not the full wind field over a large domain, but the local gusts and potential turbulence along an approach corridor and around a landing zone. These local winds can vary strongly because of terrain-induced channeling, ridge acceleration, land--sea contrast, and boundary-layer structure. Sparse new observations from UAVs or stations may also arrive immediately before the decision. The desired operating point is therefore not simply another dense high-resolution forecast, but a local wind-state estimator that can be queried continuously and corrected quickly. Fig.~\ref{fig:motivation} illustrates such a scenario.

Recent advances in AI weather forecasting and terrain-aware wind downscaling have made fast local wind estimation increasingly feasible. Large AI weather models can generate forecasts efficiently, and downscaling methods can map coarse atmospheric inputs to fine-scale terrain-dependent wind structures. However, most existing approaches still operate on fixed grids and treat downscaling as a one-shot prediction problem, which limits their ability to answer arbitrary off-grid local queries or to rapidly incorporate sparse observations after deployment. The challenge is not only one of spatial resolution. Fine-scale wind prediction over mountains and fjords cannot be obtained from low-resolution data alone: Multiple terrain-aligned high-resolution realizations can be consistent with the same kilometer-scale atmospheric state. Sparse but high-resolution observations, such as those of the drone of Fig.~\ref{fig:motivation} are therefore necessary not merely to correct small point-wise errors but, more crucially, to fully resolve local state variables.

Conditional implicit neural representations are a natural fit for this task because they define the wind field as a continuous function of coordinates, supporting off-grid queries. The main challenge lies in fast observation-guided correction. A straightforward strategy would be to fine-tune the whole network during inference after observations arrive, but this is unattractive in the sparse-observation regime: the number of observations is often far smaller than the number of trainable parameters, making overfitting likely, while repeated weight updates also increase latency and entangle reusable representation learning with case-specific state correction. Instead, we seek a formulation in which offline learning captures reusable terrain-aware structure, while online assimilation is restricted to a compact sample-specific latent state regularized by an explicit dataset-adaptive prior.

We introduce WindINR, a latent-state conditional implicit neural representation for fast local wind query and correction. Given static terrain descriptors, a low-resolution wind background, and continuous query coordinates, WindINR predicts the wind state through a shared INR decoder conditioned on a sample-specific latent state. The decoder captures reusable terrain-aware structure, while the latent state represents unresolved local variation and provides the degree of freedom for observation-guided correction. During training, a privileged reference encoder uses high-resolution supervision to define reference latents, and a deployable latent predictor estimates initial latents from inference-time inputs alone. Their discrepancies define a dataset-adaptive Gaussian prior over latent corrections. At inference time, all network weights remain fixed: without observations, WindINR decodes the initial latent state directly; with sparse observations, it updates only the low-dimensional latent state by minimizing a regularized correction objective evaluated through direct INR queries at the observation coordinates. Thus the same arbitrary-coordinate query interface is preserved before and after correction, while avoiding full-network fine-tuning.

This paper makes two main contributions:
\begin{itemize}
\item We formulate WindINR, a latent-state conditional INR for terrain-aware local wind estimation with arbitrary-coordinate continuous query.
\item We introduce a training and inference scheme for fast observation-guided latent correction, using training-set latent discrepancies to define a dataset-adaptive prior and updating only the latent state.
\end{itemize}

We train WindINR on the large-scale FuXi-CFD dataset spanning diverse terrain and evaluate it with controlled OSSEs, simulated helicopter trajectories, and real UAV measurements over \tromso{}. In OSSEs on 1,881 held-out cases, 1,024 observations reduce mean component RMSE by 13.26\%, with WindINR outperforming baselines including WindSeer. The helicopter pilot further reduces look-ahead error by 15.17\%, while real UAV experiments validate its effectiveness in mountainous terrain. By updating only a low-dimensional latent space, WindINR also achieves a 14$\times$ speedup over full-network fine-tuning while retaining arbitrary-coordinate querying.
\section{Related Work}
\label{sec:related}
This work lies at the intersection of AI weather downscaling, continuous neural representations, and observation-driven state estimation.

\paragraph{AI weather forecasting and downscaling.}
Recent AI weather models such as FourCastNet, Pangu-Weather, GraphCast, and GenCast deliver strong forecast skill and low latency, but they generally remain fixed-grid predictors \citep{fourcastnet2022,pangu2023,graphcast2023,gencast2025}. On the downscaling side, FuXi-CFD, CorrDiff, and related terrain-aware frameworks show that learned models can recover realistic fine-scale atmospheric structure from coarse inputs in complex terrain \citep{fuxicfd2026,corrdiff2025,hess2025}. WindSeer further considers observation-conditioned reconstruction, learning to infer a complete high-resolution wind field from sparse local measurements and terrain information \citep{windseer2024}; however, it likewise operates as a pretrained feed-forward model rather than performing online correction after deployment. Our focus is different: rather than full-field gridded prediction alone, we target continuous local query with fast observation-driven correction after deployment.

\paragraph{Continuous representations and implicit neural representations.}
Implicit Neural Representations were initially introduced for shape modeling \citep{deepsdf2019,occupancy2019} and image re-synthesis \citep{nerf2020} and have since been considerably refined \citep{shape2vecset2023,dora2025}. They provide a natural interface for arbitrary-coordinate and arbitrary-resolution decoding \citep{siren2020,liif2021}. In atmospheric and geophysical applications, recent works such as KANI, HyperDS, SpLIIF, and continuous field reconstruction from sparse observations show that such representations are well matched to off-grid meteorological query and flexible downscaling \citep{kani2025,hyperds2024,spliif2025,mmgn2024}. However, most existing INR-based weather methods treat the model primarily as a feed-forward downscaler or reconstructor, rather than studying how a deployed conditional INR can be updated efficiently when new sparse observations arrive.

\paragraph{Data assimilation and AI-based assimilation.}
Data assimilation combines a background estimate with observations under uncertainty assumptions to produce an updated analysis state. Recent AI-based approaches explore arbitrary-resolution neural-process assimilation, latent-space variational methods, diffusion-based DA, and forecast-assimilation systems coupled to AI weather models \citep{fnp2024,aeo2l2025,vaevar2025,lda2026,diffda2024,losda2025,appa2025,fuxida2025,hybridlatent2026}. While these methods advance learned observation-conditioned state estimation, they mostly target gridded analyses, fixed model outputs, or large generative reconstructions, rather than terrain-aware continuous local query on top of a conditional INR background.

Taken together, existing work has advanced terrain-aware downscaling, continuous atmospheric representations, and observation-driven state estimation largely along separate directions. Our work targets their intersection: a terrain-aware conditional INR for local wind estimation whose sample-specific latent state can be corrected using a dataset-adaptive prior and sparse observations, while preserving arbitrary-coordinate continuous query.

% The pre-experimental formulation and learning sections follow the supplied NIPS26.pdf.
\section{Formulation}
\label{sec:formulation}
\begin{figure}[t]
\centering
\includegraphics[width=0.95\linewidth]{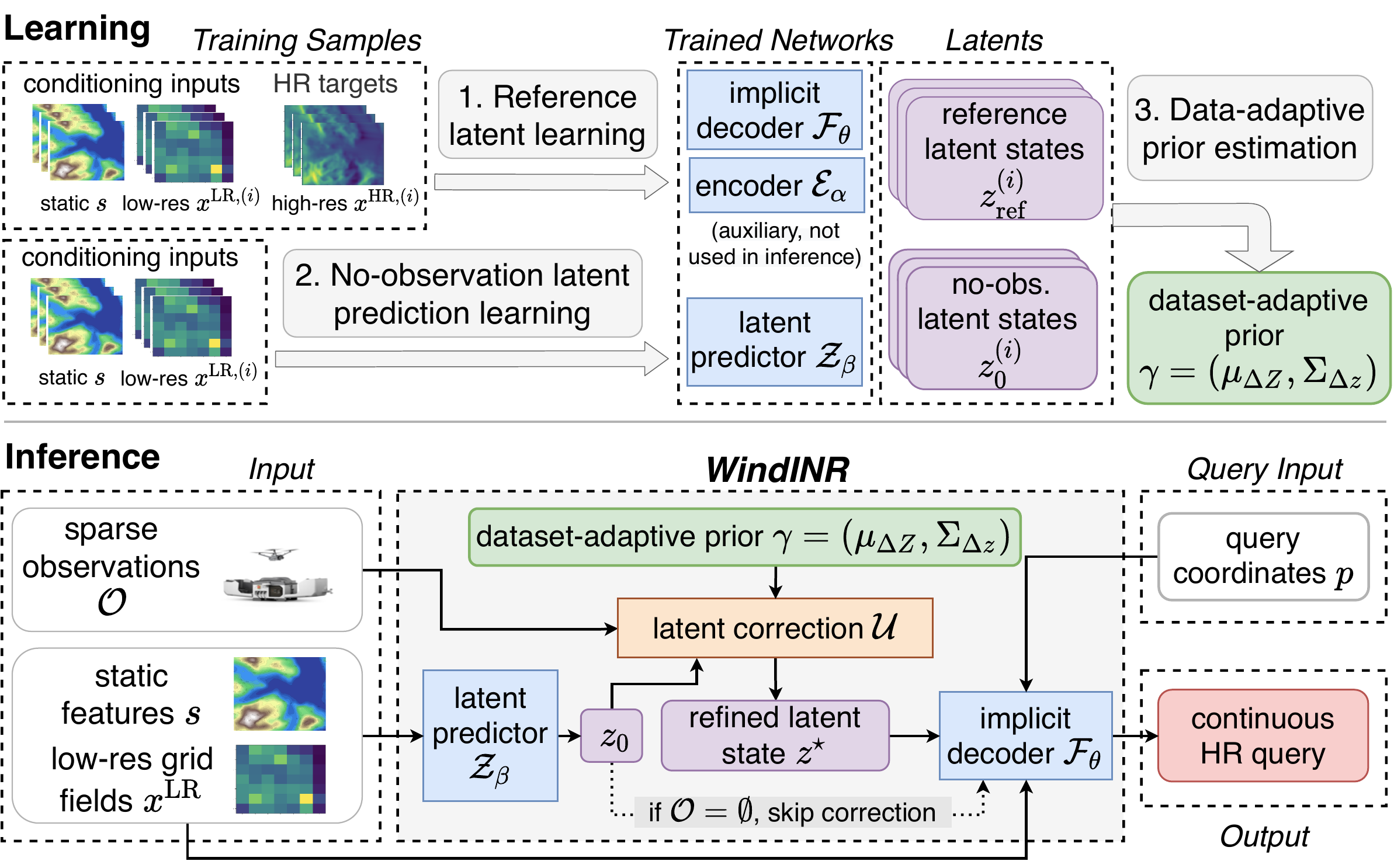}
\caption{\textbf{Overview of WindINR.} At training time, reference latent states and no-observation latent states are learned and their discrepancies are summarized into a dataset-adaptive prior $\gamma$. At inference time, sparse observations guide a regularized latent correction from the initial latent state $z_0$ to a refined latent state $z^\star$, enabling continuous high-resolution query through the implicit decoder $\mathcal{F}_\theta$.}
\label{fig:overview}
\end{figure}

Our goal is to estimate the local high-resolution wind field over complex terrain from terrain descriptors, a low-resolution background wind prediction, and sparse local observations when available. We call the resulting framework WindINR: a latent-state implicit neural representation in which a learned decoder supports continuous query at arbitrary coordinates and only a low-dimensional latent state is updated at inference time. Figure~\ref{fig:overview} summarizes the full learning-and-inference pipeline and highlights the deployed WindINR boundary at inference time.

\subsection{Latent-State INR for Querying Local Wind Fields}
\label{sec:field}
Let $\Omega\subset\R^3$ denote the continuous query domain, and let $p=(r_x,r_y,h)\in\Omega$ be a 3D query coordinate. For any given sample, that is, a target area with its static terrain descriptors $s$ and sparse observations of the local conditions $\mathcal{O}$, let $x^{\rm LR}$ be the low-resolution grid-based wind prediction, which we assume to be given \emph{a priori}, and let $x^{\rm HR}:\Omega\to\R^C$ denote the target high-resolution wind field. In our current setting, $C=3$ because we predict wind components along three axes.

A standard conditional INR can be written as
\begin{equation}
\widehat{x}^{\rm HR}(p)=\mathcal{F}_\theta\bigl(s,x^{\rm LR},p\bigr),\qquad \forall p\in\Omega,
\label{eq:standard-inr}
\end{equation}
where $\mathcal{F}_\theta$ is a decoder. This already supports arbitrary-coordinate continuous query, but the reconstruction is fully determined by the decoder weights $\theta$ and the deployable inputs $(s,x^{\rm LR})$.

To provide a sample-specific and updatable degree of freedom, we introduce a low-dimensional latent state $z\in\R^d$ and define the latent-state INR as
\begin{equation}
\widehat{x}^{\rm HR}(p;z)=\mathcal{F}_\theta\bigl(s,x^{\rm LR},p,z\bigr),\qquad p\in\Omega.
\label{eq:field}
\end{equation}
Here, $\theta$ captures reconstruction structure shared across samples, while $z$ represents sample-specific variation unresolved by $(s,x^{\rm LR})$ alone. Therefore, when new local observations become available, we refine the queried field by updating $z$ rather than modifying $\theta$. Because the backbone is an INR, predictions at observation locations are obtained by direct evaluation of $\mathcal{F}_\theta$ at the observed coordinates, so the same continuous query interface is preserved before and after latent correction.

\subsection{Inferring the Latent State}
\label{sec:latent-inference}
We now describe how the latent state is inferred at training time and at inference time.

\paragraph{Training-time reference latent state.}
During training, when high-resolution supervision is available, we define a privileged reference latent state
\begin{equation}
z_{\rm ref}=\mathcal{E}_\alpha\bigl(s,x^{\rm LR},x^{\rm HR}\bigr).
\label{eq:privileged}
\end{equation}
This latent state provides a high-quality sample-specific target in latent space during training, but it is not available at inference time.

\paragraph{Why not directly learn an observation-conditioned latent map?}
At inference time, the high-resolution target field is unavailable, so the goal is to infer a latent state that approximates the reference latent state that would have been obtained under privileged supervision. A naive strategy would be to learn a unified observation-conditioned predictor, for example $z=\mathcal{Z}_\eta\bigl(s,x^{\rm LR},\mathcal{O}\bigr)$, that directly maps the deployable inputs and the available observations to a latent state aligned with $z_{\rm ref}$. We do not adopt this fully amortized strategy because, in our setting, the observation set $\mathcal{O}$ is sparse, irregular, variable in number, and heterogeneous in quality. A single feed-forward predictor would therefore be tied to the observation layouts and noise patterns seen during training. Instead, we treat the observations at inference time as constraints that guide regularized correction of a latent estimate.

\paragraph{Two-step latent inference.}
Let $z_0=\mathcal{Z}_\beta(s,x^{\rm LR})$ denote the no-observation initial latent state. We therefore define the latent inference map as
\begin{equation}
\mathcal{Z}\bigl(s,x^{\rm LR},\mathcal{O}\bigr)=
\begin{cases}
z_0,&\mathcal{O}=\varnothing,\\
\mathcal{U}_\gamma(z_0,\mathcal{O};s,x^{\rm LR}),&\mathcal{O}\ne\varnothing.
\end{cases}
\label{eq:two-step}
\end{equation}
Here, $\mathcal{Z}_\beta$ denotes a deployable no-observation latent predictor that provides an initial latent state from $(s,x^{\rm LR})$ alone, while $\mathcal{U}_\gamma$ denotes an observation-conditioned correction operator. The concrete learning objective for $\mathcal{Z}_\beta$ and the concrete form of $\mathcal{U}_\gamma$ are introduced in Section~\ref{sec:method}. When no observations are available, $\mathcal{Z}_\beta$ returns the initial deployable latent state. When observations are available, $\mathcal{U}_\gamma$ refines this initial latent state while keeping the network weights fixed.

\paragraph{Dataset-adaptive prior for correction.}
In contrast to many previous correction schemes, where $\gamma$ is treated as a tunable regularization parameter chosen empirically, here $\gamma$ denotes the parameters of a dataset-adaptive prior for latent correction. Specifically, over the training set we consider the discrepancies between the reference codes and their matched no-observation codes, $\Delta z^{(i)}=z_{\rm ref}^{(i)}-z_0^{(i)}$, and summarize them through their empirical mean and covariance,
\(
\gamma=(\mu_{\Delta z},\Sigma_{\Delta z}).
\)
Rather than imposing a generic isotropic $\ell_2$ penalty with a hand-tuned weight, this formulation uses training-set latent discrepancies to define a more informative prior, thereby guiding the correction toward latent-error patterns supported by the data.

The final high-resolution field remains continuously queryable as
\(
\widehat{x}^{\rm HR}(p;\mathcal{O})=\mathcal{F}_\theta\bigl(s,x^{\rm LR},p,\mathcal{Z}(s,x^{\rm LR},\mathcal{O})\bigr),\qquad p\in\Omega.
\)
This separation defines the core operating principle of WindINR: the decoder learns a reusable terrain-aware representation, while sparse local observations refine only the latent state of the current local wind field.

\section{Learning and Latent Correction in WindINR}
\label{sec:method}
We now instantiate the operators that define WindINR in Section~\ref{sec:formulation}. At training time, we first learn a privileged reference latent state and then train a deployable no-observation latent predictor. We next estimate the parameters of the dataset-adaptive prior from training-set latent discrepancies. At inference time, when sparse observations are available, WindINR refines the latent state by solving a regularized correction problem while keeping all network weights fixed.

\subsection{Reference latent learning}
\label{sec:training}
We first learn the implicit decoder $\mathcal{F}_\theta$ together with the reference latent encoder $\mathcal{E}_\alpha$. Although Section~\ref{sec:latent-inference} writes the reference latent state abstractly as $z_{\rm ref}=\mathcal{E}_\alpha(s,x^{\rm LR},x^{\rm HR})$, in implementation the high-resolution target field enters through sampled support and query sets. For each training sample, we draw $\mathcal{P}^{\rm sup}=\{(p_i,x^{\rm HR}(p_i))\}_{i=1}^{m_{\rm sup}}$, $\mathcal{P}^{\rm qry}=\{(q_j,x^{\rm HR}(q_j))\}_{j=1}^{m_{\rm qry}}$, where $p_i,q_j\in\Omega$ are continuous coordinates sampled from the high-resolution domain. The reference latent state is inferred from the deployable inputs together with the support set: $z_{\rm ref}=\mathcal{E}_\alpha\bigl(s,x^{\rm LR},\mathcal{P}^{\rm sup}\bigr)$. Conditioned on this latent state, the decoder predicts the field at held-out query coordinates, $\widehat{x}^{\rm HR}_{\rm ref}(q_j)=\mathcal{F}_\theta\bigl(s,x^{\rm LR},q_j,z_{\rm ref}\bigr)$. We train the reference branch by minimizing
\(
\mathcal{L}_{\rm ref}=\frac{1}{m_{\rm qry}}\sum_{j=1}^{m_{\rm qry}}\norm{\widehat{x}^{\rm HR}_{\rm ref}(q_j)-x^{\rm HR}(q_j)}_2^2+\lambda_{\rm ref}\norm{z_{\rm ref}}_2^2.
\)
This stage serves two roles. First, it learns $\mathcal{F}_\theta$ as a latent-conditioned INR that supports arbitrary-coordinate continuous query. Second, it defines a high-quality training-time latent state $z_{\rm ref}$ that acts as the latent target for the later deployable branch.

\subsection{No-observation latent prediction}
\label{sec:background-training}
We next instantiate the $\mathcal{O}=\varnothing$ branch in Eq.~\eqref{eq:two-step} by learning the deployable predictor $\mathcal{Z}_\beta$. For each sample, the predictor returns the initial no-observation latent state $z_0=\mathcal{Z}_\beta\bigl(s,x^{\rm LR}\bigr)$. This latent state is decoded by the frozen INR as $\widehat{x}^{\rm HR}_0(q_j)=\mathcal{F}_\theta\bigl(s,x^{\rm LR},q_j,z_0\bigr)$. At this stage, $\mathcal{F}_\theta$ and $\mathcal{E}_\alpha$ are fixed, and only $\mathcal{Z}_\beta$ is optimized. The objective combines reconstruction under the frozen decoder with alignment to the reference latent state. We take it to be
\(
\mathcal{L}_\beta=\frac{1}{m_{\rm qry}}\sum_{j=1}^{m_{\rm qry}}\norm{\widehat{x}^{\rm HR}_0(q_j)-x^{\rm HR}(q_j)}_2^2+\lambda_{\rm align}\norm{z_0-z_{\rm ref}}_2^2.
\)
The first term ensures that the no-observation latent state remains useful for reconstruction. The second term encourages $\mathcal{Z}_\beta$ to approximate the privileged reference latent state from deployable inputs alone. After this stage, WindINR can already answer continuous queries without local observations through $\widehat{x}^{\rm HR}(p)=\mathcal{F}_\theta(s,x^{\rm LR},p,z_0)$.

\subsection{Estimating the dataset-adaptive prior}
\label{sec:prior}
We now instantiate the prior parameter $\gamma$ introduced in Section~\ref{sec:latent-inference}. Over a training set of $N$ samples, we form the latent discrepancies $\Delta z^{(i)}=z_{\rm ref}^{(i)}-z_0^{(i)}$, $i=1,\ldots,N$. These samples describe how the deployable no-observation latent state typically differs from the corresponding reference latent state.

We summarize these discrepancies by their empirical mean and covariance: $\mu_{\Delta z}=\frac{1}{N}\sum_{i=1}^N\Delta z^{(i)}$, $\Sigma_{\Delta z}=\frac{1}{N-1}\sum_{i=1}^N(\Delta z^{(i)}-\mu_{\Delta z})(\Delta z^{(i)}-\mu_{\Delta z})^\top$. We then define $\gamma=(\mu_{\Delta z},\Sigma_{\Delta z})$. This yields a Gaussian prior on the latent correction increment, $p(\delta z\mid\gamma)=\mathcal{N}(\mu_{\Delta z},\Sigma_{\Delta z})$, $\delta z=z-z_0$. Compared with a generic isotropic $\ell_2$ regularizer with a hand-tuned weight, this prior captures both the typical direction and the covariance structure of latent corrections supported by the training data.

\subsection{Observation-guided latent correction}
\label{sec:correction}
When local observations are available, WindINR instantiates the $\mathcal{O}\ne\varnothing$ branch in Eq.~\eqref{eq:two-step} through the correction operator $\mathcal{U}_\gamma$. For implementation, we write $\mathcal{O}=\{(p_n,y_n,M_n,R_n)\}_{n=1}^{N_{\rm obs}}$, where $p_n\in\Omega$ is the observation coordinate, $y_n$ is the observed value, $M_n$ selects the observed variables, and $R_n$ is the corresponding observation-error covariance or variance.

We implement $\mathcal{U}_\gamma$ as regularized latent correction: $\delta z^\star=\arg\min_{\delta z\in\R^d}\mathcal{J}(\delta z;z_0,\mathcal{O},\gamma)$, with
\begin{equation}
\begin{aligned}
\mathcal{J}(\delta z;z_0,\mathcal{O},\gamma)
={}&\frac12\norm{\delta z-\mu_{\Delta z}}_{(\Sigma_{\Delta z}+\varepsilon I)^{-1}}^2\\
&+\frac12\sum_{n=1}^{N_{\rm obs}}\norm{M_n\mathcal{F}_\theta\bigl(s,x^{\rm LR},p_n,z_0+\delta z\bigr)-y_n}_{R_n^{-1}}^2,
\end{aligned}
\label{eq:map}
\end{equation}
where $\varepsilon>0$ is a small numerical-stability constant and $\norm{a}_{A^{-1}}^2=a^\top A^{-1}a$. The refined latent state is then $z^\star=z_0+\delta z^\star$. Thus, the first term enforces the dataset-adaptive prior induced by training-set latent discrepancies, while the second term matches the corrected latent state to the available observations through direct decoder evaluations at the observation coordinates. All network parameters inside WindINR remain fixed during this step; only the $d$-dimensional latent state is updated in this global formulation. If no observations are available, no correction is performed and the latent state remains $z_0$. A probabilistic MAP derivation of this objective is provided in Appendix~\ref{app:derivation}.

Finally, the refined high-resolution field remains continuously queryable as $\widehat{x}^{\rm HR}_\star(p)=\mathcal{F}_\theta\bigl(s,x^{\rm LR},p,z^\star\bigr)$. Therefore, WindINR instantiates the formulation in Section~\ref{sec:formulation} by preserving the continuous-query interface of the INR while restricting inference-time correction to a low-dimensional latent correction problem.

\section{Experiments}
\label{sec:experiments-submission}
We evaluate WindINR's ability to answer local wind queries and correct them using new measurements. Our study combines controlled experiments on simulated wind fields with an evaluation against real UAV observations.

\subsection{Experimental Setup}

\paragraph{Training and inference setup.}
We train WindINR on FuXi-CFD dataset \citep{fuxicfd2026}, a large collection of terrain-resolving CFD simulations. It contains 12,532 wind-field cases spanning varied terrain, providing both the volume and the diversity of terrain--wind configurations needed to learn a reusable local wind representation. Each case covers a $9\,\mathrm{km}\times9\,\mathrm{km}$ region at 30\,m horizontal resolution. We use 8,772/1,879/1,881 cases for training/validation/testing and predict the three wind components $u,v,w$ at 27 height levels between 5 and 214.29\,m, conditioned on terrain and a coarse wind background. For WindINR, network weights remain fixed when observations arrive. On FuXi-CFD dataset, we update a $3\times3$ grid of latent controls with a training-estimated prior and eight L-BFGS outer steps. The same checkpoint, prior, and solver are used across observation counts.

\paragraph{Baselines.}
We compare with five primary baselines. \textbf{FuXi-CFD} \citep{fuxicfd2026} provides a published background-reconstruction reference without local observations. \textbf{Partial FT} performs partial fine-tuning of the latent seed and final output-head layer, while \textbf{Full FT} fine-tunes all network weights using the new observations. These two strategies test whether updating network parameters is preferable to correcting WindINR's latent state. \textbf{WindSeer} learns an end-to-end mapping from sparse observations to a high-resolution wind field. Since the original architecture does not natively incorporate a low-resolution background field, we augment our local implementation with an additional module to encode and fuse the background information. The implementation is inspired by \citet{windseer2024}, with details in Appendix~\ref{app:windseer}. \textbf{Residual IDW} uses inverse-distance weighting to interpolate observation residuals and adds the resulting correction to the background. \textbf{WindINR (Iso.)} uses an isotropic Gaussian prior for ablation.

\subsection{Simulated Observations}

For the simulated-observation experiments, we use high-resolution wind fields from the FuXi-CFD dataset as ground truth. We consider two complementary observation settings: randomly sampled observations for controlled evaluation and a UAV-aided approach scenario that mimics sequential measurements along a flight trajectory.

\begin{table}[!htbp]
\centering\small
\caption{\textbf{Random simulated observations.} Case-averaged RMSE in $\ms$ over 1,881 test cases, with 2,048 held-out query points per case. ``Raw'' is WindINR before correction; gain and wins compare the corrected result with that raw field (Appendix~\ref{app:experimental-settings}).}
\label{tab:assimilation-submission}
\begin{tabular}{lrrrrrrr}
\toprule
& Raw & \multicolumn{4}{c}{Analysis RMSE} & & \\
\cmidrule(lr){3-6}
Obs. & mean & $u$ & $v$ & $w$ & Mean & Gain & Wins \\
\midrule
% Automatically generated by scripts/rebuild_results.py.
256 & 0.4051 & 0.4959 & 0.4905 & 0.1586 & 0.3817 & 5.78\% & 83.41\% \\
512 & 0.4051 & 0.4773 & 0.4716 & 0.1514 & 0.3668 & 9.45\% & 97.24\% \\
1,024 & 0.4051 & 0.4576 & 0.4523 & 0.1442 & 0.3514 & 13.26\% & 99.68\% \\
2,048 & 0.4052 & 0.4380 & 0.4333 & 0.1379 & 0.3364 & 16.98\% & 100.00\% \\

\bottomrule
\end{tabular}
\end{table}

\paragraph{Random observations.}
For each test case, we randomly sample 256, 512, 1,024, or 2,048 locations and add independent Gaussian noise with standard deviations $(0.5,0.5,0.2)$~$\ms$ to their $u,v,w$ values. Another 2,048 locations are held out for evaluation. Notably, WindSeer has limited flexibility in handling varying numbers of observations. We therefore train separate WindSeer models for each of the four observation densities considered in our experiments. We report case-averaged RMSE for each component and the arithmetic mean of the three component RMSEs. WindINR's background reconstruction has lower mean component RMSE than FuXi-CFD (0.4051 versus 0.4141~$\ms$; Table~\ref{tab:windseer-submission}). Incorporating random observations reduces the error further at every evaluated count (Table~\ref{tab:assimilation-submission}). WindINR achieves lower mean error than Partial FT, Full FT, WindSeer, and Residual IDW at all four counts (Table~\ref{tab:windseer-submission}). Consistent gains over WindINR (Iso.) demonstrate the effectiveness of the dataset-adaptive prior. Against WindSeer, WindINR improves all three components at every observation count and wins on 91.71\% of paired cases with 1,024 observations (Appendix~\ref{app:windseer-components}). This result is particularly notable because WindSeer is trained in a fully supervised manner for observation-to-field reconstruction, whereas WindINR achieves superior performance without such supervised training, highlighting the effectiveness of its test-time latent correction mechanism.

\begin{table}[!htbp]
\centering\small
\setlength{\tabcolsep}{3pt}
\caption{\textbf{Background wind queries and random-observation correction.} Mean component RMSE in $\ms$. Rows with observations use 1,881 test cases; ``0'' denotes no local observations. Bold and underlined values mark the lowest and second-lowest errors; dashes indicate inapplicable entries.}
\label{tab:windseer-submission}
\begin{tabular}{lrrrrrrr}
\toprule
Obs. & FuXi-CFD & Partial FT & Full FT & WindSeer & Residual IDW & WindINR (Iso.) & WindINR \\
\midrule
% Generated by scripts/rebuild_windseer.py; do not edit numerals manually.
0 & \underline{0.4141} & -- & -- & -- & -- & -- & \textbf{0.4051} \\
\midrule
256 & -- & 0.3962 & 0.5047 & \underline{0.3869} & 0.3990 & 0.3886 & \textbf{0.3817} \\
512 & -- & 0.3804 & 0.4837 & 0.3783 & 0.3893 & \underline{0.3696} & \textbf{0.3668} \\
1,024 & -- & 0.3696 & 0.4361 & 0.3699 & 0.3814 & \underline{0.3574} & \textbf{0.3514} \\
2,048 & -- & 0.3634 & 0.3648 & 0.3600 & 0.3751 & \underline{0.3403} & \textbf{0.3364} \\

\bottomrule
\end{tabular}
\end{table}

\begin{figure}[!htbp]
\centering
\begin{minipage}[t]{0.40\linewidth}\centering
\textbf{(a) Terrain and sensing geometry}\\[2pt]
\includegraphics[width=\linewidth]{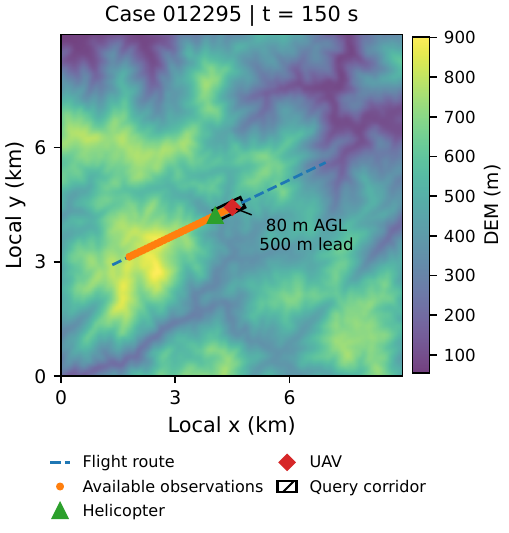}
\end{minipage}\hfill
\begin{minipage}[t]{0.58\linewidth}\centering
\textbf{(b) Error along the approach}\\[2pt]
\includegraphics[width=\linewidth]{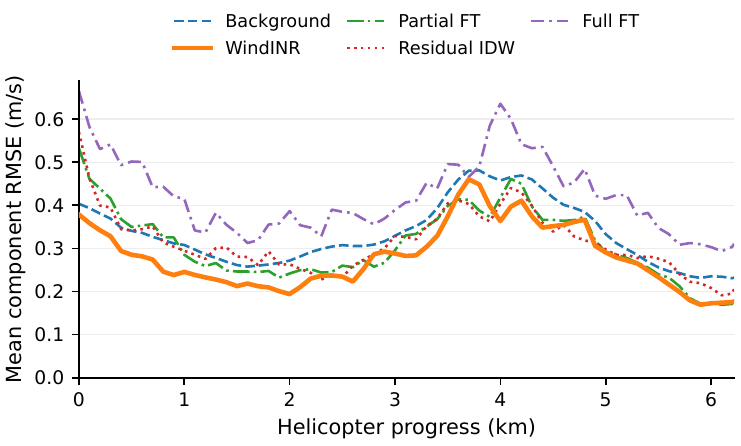}
\end{minipage}
\caption{\textbf{UAV-aided wind queries on FuXi-CFD dataset.} (a) Terrain, route, observations, and the $800\times300$\,m look-ahead query region at $t=150$\,s. (b) Mean component RMSE across eight cases at each helicopter progress point, excluding queries within 50\,m of assimilated observations. Each case queries a steady simulated wind field as the available observation prefix grows.}
\label{fig:corridor-submission}
\end{figure}

\paragraph{UAV-aided approach.}
The trajectory experiment simulates a UAV flying ahead of a helicopter along an approach route using eight held-out CFD cases. As the helicopter advances, only UAV observations already collected along the route are used to correct the estimated wind in its upcoming $800\times300$\,m corridor. We query the corrected field ahead of the helicopter and evaluate it against the CFD truth, excluding locations within 50\,m of assimilated observations. Since the CFD fields are steady, the changing observation history and moving query region test whether measurements collected earlier along the route can improve wind estimates farther ahead. Here the number of available observations changes dynamically as the UAV advances. We compare WindINR with interpolation and both fine-tuning strategies; the WindSeer models trained for fixed observation counts are not included in this dynamic-observation setting. WindINR has the lowest average error at most points along the approach (Figure~\ref{fig:corridor-submission}). WindINR reduces the background error by 15.17\% and improves all eight case averages, as shown in Table~\ref{tab:corridor-main-submission}. 

\begin{table}[!htbp]
\centering\small
\caption{\textbf{Simulated UAV-aided approach.} Query-weighted holdout RMSE in $\ms$ over eight cases, excluding queries within 50\,m of observations. ``Improved'' counts cases with lower error than the corresponding background.}
\label{tab:corridor-main-submission}
\begin{tabular}{lrrrrc}
\toprule
Method & $u$ & $v$ & $w$ & Mean & Improved \\
\midrule
Background & 0.4020 & 0.5051 & 0.1175 & 0.3415 & -- \\
WindINR & \textbf{0.3499} & \textbf{0.4117} & \textbf{0.1075} & \textbf{0.2897} & 8/8 \\
Partial FT & 0.3843 & 0.4448 & 0.1304 & 0.3198 & 6/8 \\
Residual IDW & 0.3781 & 0.4467 & 0.1444 & 0.3231 & 5/8 \\
Full FT & 0.5198 & 0.5901 & 0.1851 & 0.4317 & 1/8 \\

\bottomrule
\end{tabular}
\end{table}

\paragraph{Time efficiency.} Under the current paired UAV benchmark, WindINR reduces assimilation time from a median of $8.92$~s to $0.63$~s per evaluation point, delivering approximately 14$\times$ speedup and a 93\% reduction in runtime compared with full fine-tuning. This efficiency comes from updating only the latent state while keeping the network weights frozen, making WindINR particularly suitable for rapid online UAV-field assimilation.

\subsection{Real Observations}
\label{sec:uav}
\paragraph{Troms\o{} UAV observations and holdouts.}
We evaluate measured UAV winds collected around \tromso{}, Norway, during one episode on 4 September 2026. The coarse physical background uses the Nordic 1\,km analysis and AROME-Arctic fields; terrain and roughness are prepared separately. We consider three UAV-derived evaluation protocols: a spatial block, a survey trajectory, and a stopped transect. Each uses 16 aggregated wind measurements for correction, while 12, 19, and 24 separate measurements, respectively, are withheld as verification targets. The spatial block tests queries away from observed locations, whereas the two trajectory protocols use held-out portions of the flights. 

\paragraph{Results.}
We additionally include \textbf{Constant} as a trivial baseline. Constant adds a spatially uniform horizontal-wind offset estimated from the observations to the background at every query. WindINR outperforms Constant, Residual IDW, Partial FT, and Full FT on all three protocols (Table~\ref{tab:real-uav-all}). Overall, errors in the real-observation experiments are larger than in the simulated settings, reflecting the additional uncertainties and biases present in both the analysis background and UAV measurements. Notably, full fine-tuning becomes substantially less stable under these conditions and can even amplify the error by several times in some cases. In contrast, both Partial FT and WindINR remain considerably more robust, with WindINR consistently achieving better performance than Partial FT. These results suggest that latent-space correction provides a more reliable and effective mechanism for adapting to imperfect real-world observations than direct network fine-tuning. This robustness is particularly encouraging for extending WindINR to more challenging UAV- and helicopter-assisted operations in complex terrain, including potential search-and-rescue applications.

\begin{table}[!htbp]
\centering\small
\setlength{\tabcolsep}{3pt}
\caption{\textbf{Real-UAV horizontal-wind correction.} Horizontal-vector RMSE in $\ms$. Bold and underlined values mark the lowest and second-lowest errors among the five displayed estimators. Gain is WindINR's relative improvement over the underlined result (Appendix~\ref{app:real-uav-protocol}).}
\label{tab:real-uav-all}
\begin{tabular}{lrrrrrrr}
\toprule
Protocol & $N_h$ & Constant & Residual IDW & Partial FT & Full FT & WindINR & Gain (\%) \\
\midrule
% Generated from data/real_uav by scripts/rebuild_real_uav.py
\tromso{} spatial block & 12 & \underline{1.6365} & 1.7117 & 1.7389 & 2.8302 & \textbf{1.3773} & +15.84 \\
\tromso{} survey & 19 & 5.1026 & 5.1278 & \underline{5.0533} & 5.1923 & \textbf{5.0197} & +0.67 \\
\tromso{} stopped transect & 24 & \underline{2.8090} & 3.8671 & 2.8396 & 6.4746 & \textbf{2.2245} & +20.81 \\

\bottomrule
\end{tabular}
\end{table}
\FloatBarrier

\section{Conclusion}
\label{sec:conclusion}
WindINR provides continuous local wind estimates by separating reusable terrain-aware representation learning from sample-specific latent-state correction. A shared INR decoder supports arbitrary-coordinate queries, while a deployable predictor initializes compact latent controls from terrain and coarse background winds. Training-time discrepancies between predicted and privileged reference latents define a dataset-adaptive Gaussian prior. At inference, sparse observations update only the latent state, keeping network weights fixed and preserving the continuous query interface. Controlled OSSEs and real UAV observations support this formulation for local wind reconstruction. Future work will assess broader generalization, correlated errors, and end-to-end latency.

\label{main-text-end}

\clearpage
\section*{AI use statement}
Generative AI tools assisted with manuscript editing, LaTeX format conversion, preparation of plotting and tabulation scripts, and consistency checks and paired-case audits of the supplied experiment reports. We have reviewed all AI-assisted work. LLM-generated code was verified and tested for correctness by 2 authors. We take responsibility for the final content of this work, including text, claims or artifacts produced with the aid of generative AI.

\section*{Ethics statement}
The experiments combine simulated wind fields and synthetic observations with a small measured-UAV evaluation. Wind estimates intended for aviation or other safety-critical applications require independent validation, uncertainty assessment, and appropriate operational safeguards; the present reconstruction scores are not a safety certification. 

\section*{Reproducibility statement}
The formulation and MAP interpretation are given in Section~\ref{sec:method} and Appendix~\ref{app:derivation}. Appendix~\ref{app:repro} records the case split, selection rule, model configuration, inference settings, and the scope of available evidence. Appendix~\ref{app:windseer} adds the density-specific supervised baseline, checkpoint-selection record, and paired evaluation checks. The accompanying source package includes anonymized per-case metric records, a case manifest, and scripts that regenerate the tables and plots. Appendix~\ref{app:corridor} adds the mobile protocol, full per-time metric records, terrain arrays, and the distinction between update-time measurements and end-to-end latency. Appendix~\ref{app:real-uav} records the real-flight protocol roles, the reported evaluation subset, and the limits of the available point-level record. The main real-UAV table is regenerated from the corresponding supplied summaries; no new model inference was performed for this revision. 

\bibliography{references}
\bibliographystyle{iclr2027_conference}
\clearpage
\appendix
\section{MAP derivation and the influence of sparse observations}
\label{app:derivation}
For the notation below, $b\equiv x^{\rm LR}$, $x(p)\equiv x^{\rm HR}(p)$, and $\mu\equiv\mu_{\Delta z}$. The regularized prior covariance is denoted by $B_z$; the diagonal, scaled covariance used in the FuXi-CFD experiments is specified in Appendix~\ref{app:repro}.
\subsection{Gaussian assumptions and the correction objective}
Consider one case, with fixed terrain $s$, background input $b$, and deployable latent $z_0$. Assume a latent correction prior $\delta z\sim\mathcal{N}(\mu,B_z)$ with $B_z\succ0$. Conditional on the latent state, assume observation errors are independent with known covariances $R_n\succ0$. Then
\begin{align}
 p(\delta z\mid\mathcal{O},s,b,z_0)
 &\propto p(\delta z)\prod_n p(y_n\mid\delta z,s,b,z_0),\\
 p(y_n\mid\delta z,s,b,z_0)
 &=\mathcal{N}\!\left(M_n\mathcal{F}_{\theta}(s,b,p_n,z_0+\delta z),R_n\right).
\end{align}
Taking the negative log and discarding constants independent of $\delta z$ gives Eq.~\eqref{eq:map}. The quadratic notation is $\norm{r}_{A}^2=r^\top A r$. If only a subset of wind components is observed, $M_n$ and $R_n$ are restricted to those components. Correlated errors can instead be represented by a joint observation covariance; the independent-noise likelihood is an explicit assumption of the present OSSE.

A correction prior on $\delta z$ is different from assuming $z\sim\mathcal{N}(0,I)$. The reference-to-background discrepancies characterize errors of the deployable latent predictor in the decoder's learned coordinate system. Their nonzero mean describes a typical correction, and their variances describe relative scales on which latent coordinates may change. 

\subsection{Whitening and spatial controls}
In the global formulation, let $B_z=LL^\top$ and set $\delta z=\mu+La$. Equation~\eqref{eq:map} becomes
\begin{equation}
 \min_a\;\frac12\norm{a}_2^2+
 \frac12\sum_n\norm{M_n\mathcal{F}_{\theta}(s,b,p_n,z_0+\mu+La)-y_n}_{R_n^{-1}}^2.
 \label{eq:whitened}
\end{equation}
Whitening absorbs the prior covariance into the decoder's latent input and avoids explicitly applying an inverse covariance at every objective evaluation. With a diagonal covariance this is a component-wise rescaling, not a rotation learned from cross-coordinate covariance.

The same construction applies to spatial controls. Let $\eta$ collect coefficients on a control grid, let $\delta z(p;\eta)$ denote their fixed spatial readout, and let their quadratic prior have mean $\bar\eta$ and covariance $C_\eta$. Define $\eta=\bar\eta+L_\eta a$. The observation operator then becomes
\begin{equation}
 h(a)=\operatorname{concat}_{n}\left[
 M_n\mathcal{F}_{\theta}\bigl(s,b,p_n,z_0+\delta z(p_n;\bar\eta+L_\eta a)\bigr)
 \right].
\end{equation}

\subsection{A local linear view}
Stack the observations into $y$ and their covariances into $R$. Around the whitened prior center, linearize the observation operator as $h(a)\approx h(0)+Ga$, where $G=\partial h/\partial a\vert_{a=0}$. The local quadratic problem is
\begin{align}
 &\min_a\;\frac12 a^\top a+\frac12\norm{Ga-d}_{R^{-1}}^2,\qquad d=y-h(0),\notag\\
 &a^*=(I+G^\top R^{-1}G)^{-1}G^\top R^{-1}d.
 \label{eq:linear-map}
\end{align}
For a new query point, let $G_q$ be the Jacobian of the decoded wind with respect to the same controls. Its first-order analysis increment is
\begin{equation}
 \widehat{x}_{\rm an}(q)-\widehat{x}_{\rm prior}(q)
 \approx G_q(I+G^\top R^{-1}G)^{-1}G^\top R^{-1}d.
 \label{eq:influence}
\end{equation}
Thus observations can change an unobserved query through shared latent sensitivity, not only through geographic proximity. The learned representation determines which field changes are expressible, while the innovation $d$ determines which are supported by the current observations. Directions invisible to the observations remain controlled by the prior.

\section{FuXi-CFD implementation and evaluation record}
\label{app:repro}
\subsection{Dataset, split, and metric definitions}
\label{app:experimental-settings}
\paragraph{Simulation data.}
FuXi-CFD provides terrain-resolving CFD wind fields over $9\,\mathrm{km}\times9\,\mathrm{km}$ regions at 30\,m horizontal resolution \citep{fuxicfd2026}. Our processed collection contains 12,532 cases, split into 8,772/1,879/1,881 training/validation/test cases. We reconstruct $u,v,w$, not turbulent kinetic energy. The exported evaluation covers 27 AGL levels from 5 to 214.29\,m, not the entire vertical CFD domain. AGL is height above local ground and is distinct from altitude above sea level.
The case identifiers are unique and disjoint across the three subsets; the split seed is 42.

The 27 evaluated AGL levels, rounded to two decimals, are
\begin{quote}
$5,10,15,20,25,30,35,40,45,50,55,60,65,70,75,80,85,90,95,100,$\\
$106.50,114.95,125.94,140.22,158.78,182.91,214.29$ metres.
\end{quote}

\paragraph{Metrics.}
For the observation-density evaluation, each test case has $N_h=2,048$ holdout points that are reserved for evaluation rather than used in the observation loss. For component $c$, we report
\begin{equation}
 e_{i,c}=\sqrt{\frac{1}{N_h}\sum_{j=1}^{N_h}
   (\widehat{x}_{i,c}(q_j)-x_{i,c}(q_j))^2},\qquad
 E_c=\frac{1}{N_{\rm test}}\sum_i e_{i,c},\quad
 \overline{E}=\frac{E_u+E_v+E_w}{3}.
 \label{eq:metrics}
\end{equation}
All errors are in $\ms$. Thus $\overline{E}$ is a mean of component RMSEs, \emph{not} a vector RMSE or an RMSE pooled over every point. Relative improvement is $100(\overline{E}_{\rm bg}-\overline{E}_{\rm an})/\overline{E}_{\rm bg}$. The case win rate counts strict reductions of the corresponding per-case component mean. Test statistics are recomputed from the per-case records.
All primary test means are computed over the 1,881 exported cases, each with 2,048 evaluation points, for a total of 3,852,288 query points per observation density.

\FloatBarrier

\subsection{WindINR training and model selection}
\paragraph{Model selection and observations.}
The final WindINR checkpoint is chosen by $w$ RMSE on a fixed 256-case validation subset with 2,048 query points per case. The latest refinement uses four A100 GPUs at most and leaves the 3.55M-parameter architecture unchanged. The test evaluation follows freezing of the selected checkpoint. Synthetic observations sample the CFD field with independent Gaussian noise of standard deviation $(0.5,0.5,0.2)$~$\ms$ for $(u,v,w)$. All WindINR observation-density runs use this frozen model, the re-estimated prior, and the same online solver configuration. The supervised comparator is separately trained on the same training split, as detailed in Section~\ref{sec:experiments-submission}.

\begin{table}[htbp]
\centering\small
\caption{WindINR training configuration.}
\label{tab:config}
\begin{tabular}{ll}
\toprule
Setting & Value \\
\midrule
Terrain / coarse input channels & 4 / 3 \\
Terrain / coarse context feature widths & 64 / 32 \\
Latent channel dimension & 128 \\
Point encoder / decoder / background widths & 256 / 256 / 256 \\
Decoder layers / dropout & 4 / 0 \\
Reference pooling width & 128 \\
Positional encoding frequencies & 10 \\
Multi-scale spatial context & Enabled \\
Pyramid query width & 32 \\
Background readout / grid & Spatial / $4\times4$ \\
Baseline mode & Constant \\
Reported total / Stage~1 trainable parameters & 3,546,308 / 3,546,308 \\
Stage~2 trainable parameters & 990,432 \\
Support / query samples per case & 8,192 / 8,192 \\
Training batch size & 16 cases \\
Final component loss weights & $1:1:2$ for $u:v:w$ \\
Stage~2 latent alignment weight & 1.0 \\
Maximum concurrent GPUs & 4 NVIDIA A100-PCIE-40GB \\
\bottomrule
\end{tabular}
\end{table}
\FloatBarrier

\subsection{Latent-prior estimation and online settings}
The prior is re-estimated after the final model is frozen. Its fit split is the 8,772-case training set, using 8,192 high-resolution support points per case; the separate 1,879-case validation set supplies diagnostics only. The export records a shrinkage coefficient of 0.1, numerical regularization $10^{-4}$, and 128 latent channels. 

\begin{table}[htbp]
\centering\small
\caption{Observation-density evaluation settings.}
\begin{tabular}{ll}
\toprule
Setting & Value \\
\midrule
Observed components & $u,v,w$ \\
Observation-noise standard deviation & $(0.5,0.5,0.2)$~$\ms$ \\
Observation counts per case & 256, 512, 1,024, 2,048 \\
Independent evaluation points per case & 2,048 \\
Prior mode / covariance type & Adaptive / diagonal \\
Covariance scale / whitening & 64 / enabled \\
Online spatial latent grid & $3\times3$ \\
Optimizer / outer steps & L-BFGS / 8 \\
Learning rate / gradient clipping & 1 / 10 \\
Network-weight updates & None \\
\bottomrule
\end{tabular}
\end{table}
\FloatBarrier

\subsection{Background reconstruction}
\label{app:background-reconstruction}
Table~\ref{tab:background-appendix} gives the component-wise background reconstruction results without local observations. WindINR achieves lower mean and horizontal-component RMSE than FuXi-CFD, while FuXi-CFD has lower vertical-component error. 

\begin{table}[htbp]
\centering\small
\caption{\textbf{Background reconstruction without local observations.} Component RMSE and their arithmetic mean in $\ms$. FuXi-CFD values are from Supplementary Table S1 of \citet{fuxicfd2026}, using frequency-domain loss.}
\label{tab:background-appendix}
\begin{tabular}{lrrrr}
\toprule
Model & $u$ & $v$ & $w$ & Mean \\
\midrule
% Automatically generated by scripts/rebuild_results.py.
FuXi-CFD & 0.5486 & 0.5473 & 0.1465 & 0.4141 \\
WindINR & 0.5216 & 0.5234 & 0.1704 & 0.4051 \\

\bottomrule
\end{tabular}
\end{table}
\FloatBarrier

\subsection{Density-matched supervised baseline: implementation and training}
\label{app:windseer}

\paragraph{Offline optimization.}
All four baseline models use the same 8,772 training cases and 1,879 validation cases as the shared FuXi-CFD split. Each uses seed 0, 16 cases per batch, 8,192 query points per case, a maximum of 20 epochs, learning rate $2\times10^{-4}$, and weight decay $10^{-4}$. Configuration values are hidden width 256, observation width 128, four layers, eight positional-encoding frequencies, zero dropout, IDW power 2, and IDW epsilon $10^{-6}$. The raw Stage~2 WindINR background is frozen. The observation count matches the intended test density, and noise standard deviations are $(0.5,0.5,0.2)$~$\ms$.

Early stopping uses validation \emph{loss}, patience 5, and minimum improvement $10^{-4}$. Table~\ref{tab:windseer-training} records the last saved checkpoint rather than selecting a later epoch by validation RMSE or test performance. Baseline validation uses the full validation split, whereas WindINR's last refinement is selected using the fixed 256-case subset described above. Both estimators use the same frozen background.

\begin{table}[htbp]
\centering\small
\caption{Supervised-baseline training record. Loss is the exported validation objective; it is not assigned physical RMSE units. }
\label{tab:windseer-training}
\begin{tabular}{rrrrr}
\toprule
Observations & Epochs completed & Selected epoch & Val. loss & Val. mean RMSE \\
\midrule
% Generated by scripts/rebuild_windseer.py; do not edit numerals manually.
256 & 18 & 13 & 0.029364 & 0.4326 \\
512 & 20 & 16 & 0.028093 & 0.4224 \\
1,024 & 19 & 14 & 0.026885 & 0.4124 \\
2,048 & 17 & 12 & 0.025369 & 0.4012 \\

\bottomrule
\end{tabular}
\end{table}

\FloatBarrier
\subsection{Component errors and case-wise comparisons}
\label{app:windseer-components}
\label{app:experiments}
\begin{table}[htbp]
\centering\small
\caption{Component-wise comparison on the paired FuXi-CFD test evaluations. RMSE and MAE are in $\ms$ and averaged over cases. Bold RMSE values are lower within each density pair. ``Mean MAE'' is the arithmetic mean over $u,v,w$.}
\label{tab:windseer-components}
\begin{tabular}{llrrrrr}
\toprule
& & \multicolumn{4}{c}{RMSE} & \\
\cmidrule(lr){3-6}
Obs. & Method & $u$ & $v$ & $w$ & Mean & Mean MAE \\
\midrule
% Generated by scripts/rebuild_windseer.py; do not edit numerals manually.
256 & WindSeer & 0.5001 & 0.4991 & 0.1615 & 0.3869 & 0.2640 \\
256 & WindINR & \textbf{0.4959} & \textbf{0.4905} & \textbf{0.1586} & \textbf{0.3817} & 0.2598 \\
\addlinespace[3pt]
512 & WindSeer & 0.4882 & 0.4877 & 0.1588 & 0.3783 & 0.2574 \\
512 & WindINR & \textbf{0.4773} & \textbf{0.4716} & \textbf{0.1514} & \textbf{0.3668} & 0.2502 \\
\addlinespace[3pt]
1,024 & WindSeer & 0.4799 & 0.4744 & 0.1554 & 0.3699 & 0.2521 \\
1,024 & WindINR & \textbf{0.4576} & \textbf{0.4523} & \textbf{0.1442} & \textbf{0.3514} & 0.2406 \\
\addlinespace[3pt]
2,048 & WindSeer & 0.4669 & 0.4615 & 0.1515 & 0.3600 & 0.2449 \\
2,048 & WindINR & \textbf{0.4380} & \textbf{0.4333} & \textbf{0.1379} & \textbf{0.3364} & 0.2313 \\

\bottomrule
\end{tabular}
\end{table}

The aggregate advantage is present in all three components. At 1,024 observations, the local supervised baseline has $u/v/w$ RMSE $0.4799/0.4744/0.1554$, compared with $0.4576/0.4523/0.1442$ for WindINR. At 2,048 observations the corresponding values are $0.4669/0.4615/0.1515$ and $0.4380/0.4333/0.1379$. These are averages over the same cases, not claims that every query point or case improves.

\begin{figure}[!htbp]
\centering
\begin{minipage}[t]{0.49\linewidth}\centering
\includegraphics[width=\linewidth]{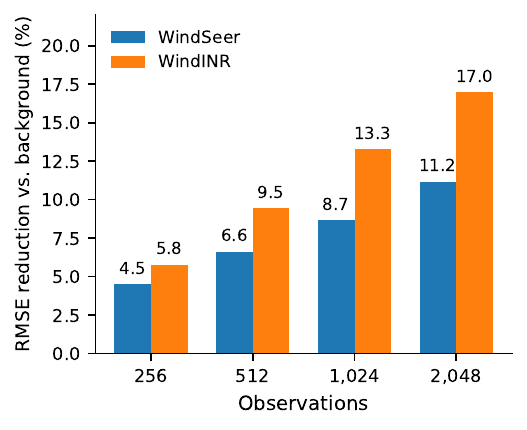}
\end{minipage}\hfill
\begin{minipage}[t]{0.49\linewidth}\centering
\includegraphics[width=\linewidth]{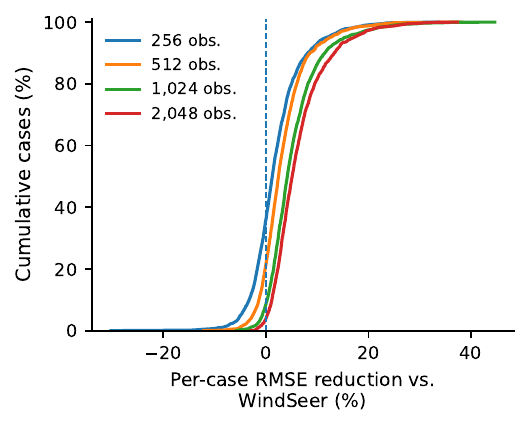}
\end{minipage}
\caption{\textbf{Same background, two observation-conditioned estimators.} Left: relative reduction of mean component RMSE against the corresponding background. Right: empirical distributions of paired per-case relative RMSE reductions, using each case's WindSeer error as denominator; positive values favor WindINR and negative values favor the supervised baseline. All 1,881 cases are included at each density. These curves describe case variability, not uncertainty over training seeds.}
\label{fig:density}
\end{figure}

For case $i$, define $d_i=\overline e_{i,\mathrm{sup}}-\overline e_{i,\mathrm{INR}}$, where each $\overline e_i$ is the mean of the case's three component RMSEs. Figure~\ref{fig:density} plots the empirical distribution of the relative reduction $100d_i/\overline e_{i,\mathrm{sup}}$, and the paired win rate is $N^{-1}\sum_i\mathbf{1}[d_i>0]$. Rankings in Table~\ref{tab:windseer-submission} use values before rounding. The paired curves and win rates retain WindSeer as their comparator. In contrast, reduction relative to the background compares each method with its own raw score. Table~\ref{tab:windseer-wins} keeps these two notions distinct. At 256 observations, the supervised estimator improves more cases relative to the raw background (89.42\% versus 83.41\%), even though WindINR has the lower mean error and wins the direct paired comparison on 64.33\% of cases. The two robustness summaries need not rank methods identically.

\begin{table}[htbp]
\centering\small
\caption{Case-wise strict improvement rates. The middle columns compare each estimator with the shared raw background; the last column compares the two estimators directly. No ties count as wins.}
\label{tab:windseer-wins}
\begin{tabular}{lrrr}
\toprule
Obs. & WindSeer vs. raw & WindINR vs. raw & WindINR vs. supervised \\
\midrule
% Generated by scripts/rebuild_windseer.py; do not edit numerals manually.
256 & 89.42\% & 83.41\% & 64.33\% \\
512 & 95.16\% & 97.24\% & 78.95\% \\
1,024 & 94.47\% & 99.68\% & 91.71\% \\
2,048 & 97.98\% & 100.00\% & 96.28\% \\

\bottomrule
\end{tabular}
\end{table}

\FloatBarrier

The distributions retain cases where supervised correction is better: WindINR wins on 64.33\% of cases at the lowest density and 96.28\% at the highest. Mean superiority therefore does not imply a pointwise or per-case guarantee. The comparison concerns the supplied local baseline and shared-background protocol.
\FloatBarrier

\clearpage
\section{FuXi-CFD mobile-observation experiment: protocols and complete results}
\label{app:corridor}
This appendix details the simulated UAV-ahead-of-helicopter evaluation in Section~\ref{sec:experiments-submission}. It uses the frozen model from the large FuXi-CFD experiment, not a model trained separately for a selected region. Synthetic observations and verification come from the same steady CFD field. The measured-flight experiments remain separate in Appendix~\ref{app:real-uav}.

\subsection{Case selection, checkpoint, and coordinate system}
The eight cases are entries 0--7 of the existing 1,881-case test manifest (split seed 42). Their ordered identifiers match the manuscript's test manifest. Selection uses manifest order, not error or improvement. The supplied report states that the route and baseline settings were not tuned on these cases. This is a pilot subset, not a second evaluation of all 1,881 cases or a verified terrain-disjoint split.

The exported DEM and aerodynamic roughness arrays have shape $300\times300$ with $[y,x]$ indexing. Local coordinates are $x_j=y_j=30j$ metres for $j=0,\ldots,299$, spanning 0--8,970\,m. The source-input inventory records \texttt{dem}, \texttt{roughness}, \texttt{u\_100m}, and \texttt{v\_100m}. These field names do not imply 100\,m horizontal grid spacing, nor do they specify the model's complete derived-channel preprocessing. 

\begin{figure}[!ht]
\centering
\includegraphics[width=.40\linewidth]{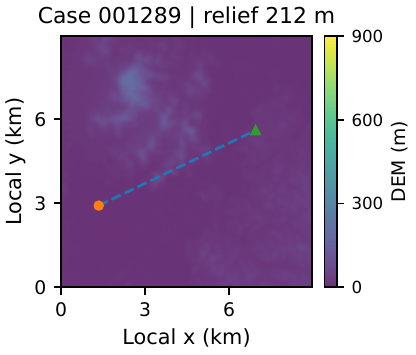}\hfill
\includegraphics[width=.40\linewidth]{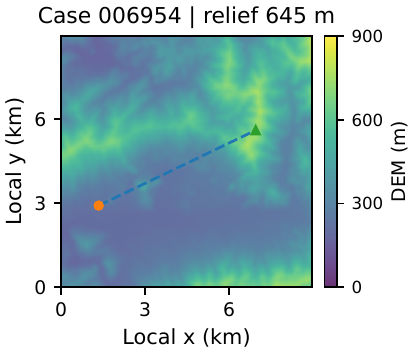}\\[3pt]
\includegraphics[width=.40\linewidth]{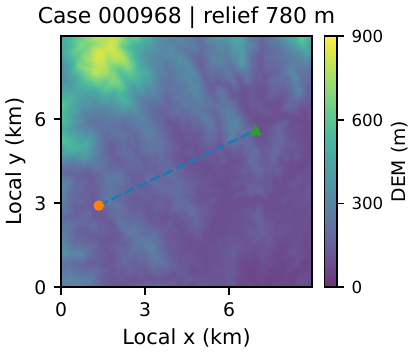}\hfill
\includegraphics[width=.40\linewidth]{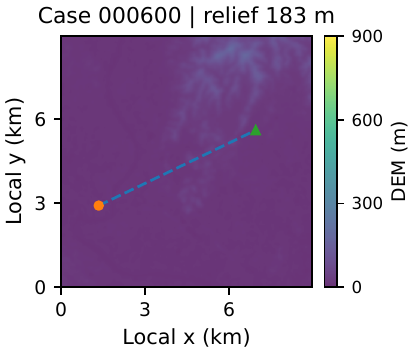}\\[3pt]
\includegraphics[width=.40\linewidth]{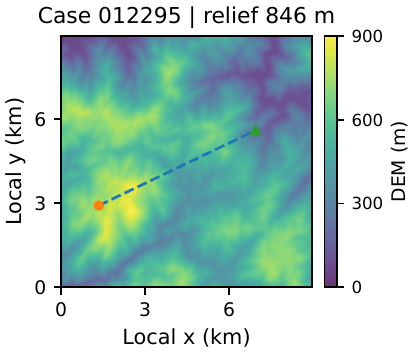}\hfill
\includegraphics[width=.40\linewidth]{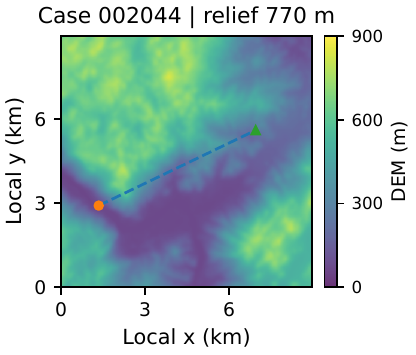}\\[3pt]
\includegraphics[width=.40\linewidth]{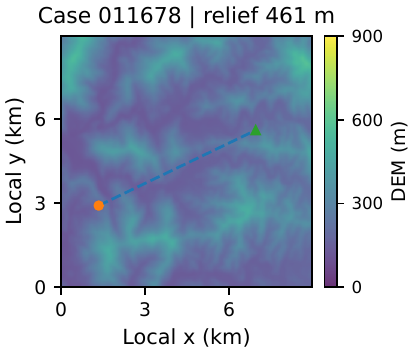}\hfill
\includegraphics[width=.40\linewidth]{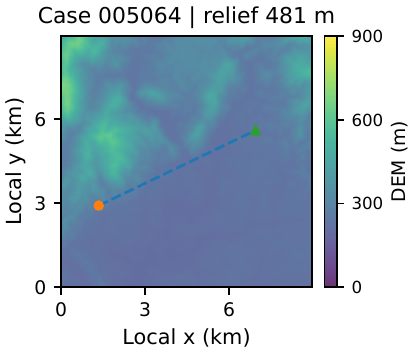}
\caption{\textbf{The eight FuXi-CFD terrain arrays.} DEM values use a shared color scale; axes are local kilometres. Dashed lines show the same normalized flight route in each case; circles and triangles mark its start and end. Terrain relief is the array maximum minus minimum.}
\label{fig:corridor-terrains}
\end{figure}

Table~\ref{tab:corridor-terrain} summarizes the terrain context for the eight cases; the corresponding maps are shown in Figure~\ref{fig:corridor-terrains}.

\begin{table}[htbp]
\centering\small
\caption{Terrain context recomputed from the exported arrays. DEM statistics are input values; they are not inferred wind fields or geographic-position metadata.}
\label{tab:corridor-terrain}
\begin{tabular}{lrrrr}
\toprule
Case ID & DEM min (m) & DEM max (m) & Relief (m) & Mean roughness (m) \\
\midrule
\texttt{001289} & 0.0 & 212.0 & 212.0 & 0.685 \\
\texttt{006954} & 177.0 & 821.9 & 644.9 & 0.837 \\
\texttt{000968} & 56.2 & 835.8 & 779.7 & 0.905 \\
\texttt{000600} & 1.5 & 184.2 & 182.8 & 0.526 \\
\texttt{012295} & 54.0 & 900.4 & 846.4 & 0.945 \\
\texttt{002044} & 63.0 & 832.6 & 769.6 & 0.789 \\
\texttt{011678} & 115.0 & 575.6 & 460.6 & 0.972 \\
\texttt{005064} & 198.0 & 678.6 & 480.6 & 0.621 \\

\bottomrule
\end{tabular}
\end{table}
\FloatBarrier

\subsection{Causal sensing and a moving verification region}
The route endpoints are $(-0.70,-0.35)$ and $(0.55,0.25)$ in normalized model coordinates. In local metres they are $(1345.50,2915.25)$ and $(6951.75,5606.25)$, giving a 6,218.64\,m route. The helicopter and UAV follow the same centerline at 80\,m AGL and 20~$\ms$, with the UAV 500\,m ahead until the endpoint clips its progress. Thus the UAV does not remain 500\,m ahead after reaching the route end. 

The first UAV observation is at 500\,m along the route at helicopter time zero. Further observations arrive every 100\,m, with an additional endpoint sample, for 59 observations in total. Each includes $(u,v,w)$ with independent Gaussian noise of standard deviation $(0.5,0.5,0.2)$~$\ms$; the base noise seed is 2026. The 64 helicopter evaluations occur at 100\,m intervals and the endpoint, covering 0--310.93\,s. Only observations with timestamps no later than the current evaluation are available. The observation prefix grows, but every analysis restarts from the background: neither latent controls nor fine-tuned weights are carried forward.

The look-ahead footprint spans 0--800\,m along track and $\pm150$\,m across track, with nominal sampling $16\times7$ at 80\,m AGL. Queries beyond the route endpoint are removed; the full query count consequently falls from 112 to 7 near the end. Verification truth is obtained by trilinear interpolation on the native $(\mathrm{AGL},y,x)$ CFD output grid. High-resolution truth is used for synthetic sensing and verification, not as a deployable model input.

The primary score excludes queries within 50\,m of any assimilated observation in $(x,y,\mathrm{AGL})$ coordinates. All methods have the same retained count at each case and time. There are 52,640 full-corridor and 47,152 retained query--time evaluations across the eight cases; the retained count ranges from 6 to 110 per record. These are not counts of independent wind realizations or necessarily distinct physical positions. Adjacent corridors overlap, and every time row for a case samples one stationary CFD field.

\subsection{Comparator recipes and scope of the comparison}
WindINR uses the unchanged $3\times3$ spatial latent control, adaptive diagonal prior, covariance scale 64, whitening, and L-BFGS with eight outer steps, learning rate 1, and gradient clipping 10. Partial FT uses Adam for 100 steps at learning rate $10^{-4}$, updating only \texttt{latent\_seed} and the final output-head layer with anchor weight 0.01. Full FT uses Adam for 100 steps at $10^{-4}$ on the full network. Both weight-update recipes specify zero weight decay and no gradient clipping. Residual IDW interpolates observation residuals relative to the same bias-corrected background with inverse-distance exponent 2.

\subsection{Aggregation and complete numerical comparison}
Let $e_{itmc}$ be the component RMSE for case $i$, evaluation time $t$, method $m$, and component $c$, and let $n_{it}$ be the query count for the chosen evaluation set. The reported mobile score is
\begin{equation}
 E_{mc}^{\rm mobile}=\frac{\sum_{i,t}n_{it}e_{itmc}}{\sum_{i,t}n_{it}},\qquad
 \overline E_m^{\rm mobile}=\frac{1}{3}\sum_{c\in\{u,v,w\}}E_{mc}^{\rm mobile}.
 \label{eq:mobile-metric}
\end{equation}
This is a query-count-weighted mean of per-record RMSEs, not a pooled pointwise RMSE. Full-corridor and holdout scores use their own counts. Every case has the same geometry and total retained count, so averaging the eight query-weighted case scores yields the same aggregate. A case is improved when its weighted mean component RMSE is below its background score; time-step win counts are descriptive only, since the times are correlated.

\begin{table}[htbp]
\centering\small
\caption{\textbf{Full-corridor sensitivity evaluation.} Component RMSE, mean component RMSE, and mean component MAE in $\ms$. These scores include nearby queries removed by the primary 50\,m exclusion. Primary holdout RMSE is reported only in Table~\ref{tab:corridor-main-submission}.}
\label{tab:corridor-full}
\begin{tabular}{llrrrrr}
\toprule
Evaluation & Method & $u$ & $v$ & $w$ & Mean & MAE \\
\midrule
Full corridor & Background & 0.4019 & 0.5058 & 0.1171 & 0.3416 & 0.2930 \\
Full corridor & WindINR & 0.3438 & 0.4024 & 0.1054 & 0.2839 & 0.2355 \\
Full corridor & Partial FT & 0.3756 & 0.4315 & 0.1272 & 0.3114 & 0.2560 \\
Full corridor & Residual IDW & 0.3860 & 0.4471 & 0.1494 & 0.3275 & 0.2668 \\
Full corridor & Full FT & 0.5131 & 0.5760 & 0.1831 & 0.4241 & 0.3410 \\

\bottomrule
\end{tabular}
\end{table}

The holdout reductions relative to the bias-corrected background are 15.17\% for WindINR, 6.35\% for Partial FT, and 5.41\% for Residual IDW; Full FT increases error by 26.40\%. WindINR reduces $u/v/w$ RMSE by 12.94/18.49/8.48\%, whereas Partial FT and Residual IDW worsen the vertical component. The aggregate ordering is unchanged when nearby queries are restored (Table~\ref{tab:corridor-full}), but this does not remove the need for the exclusion test.

\clearpage
\subsection{Along-route error for every case}
\begin{figure}[ht]
\centering
\includegraphics[width=\linewidth]{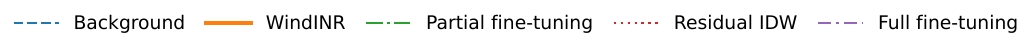}\\[4pt]
\includegraphics[width=.48\linewidth]{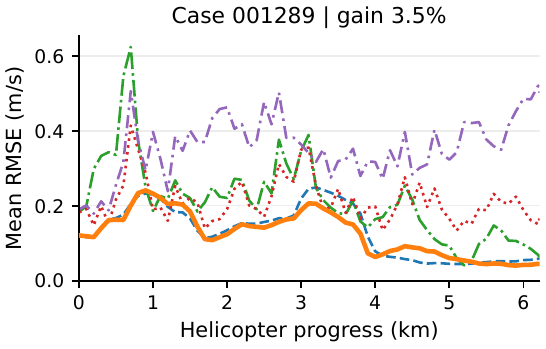}\hfill
\includegraphics[width=.48\linewidth]{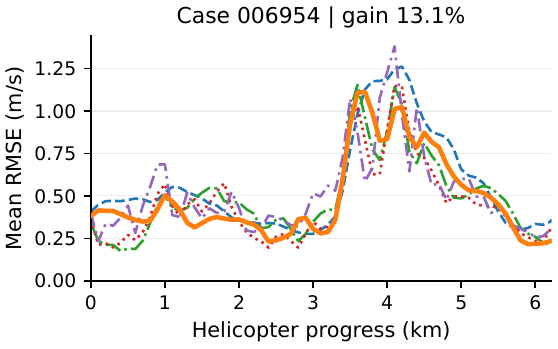}\\[4pt]
\includegraphics[width=.48\linewidth]{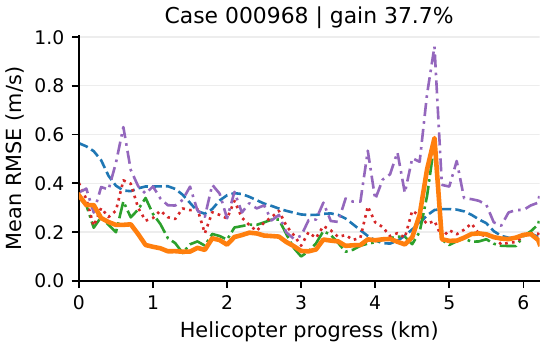}\hfill
\includegraphics[width=.48\linewidth]{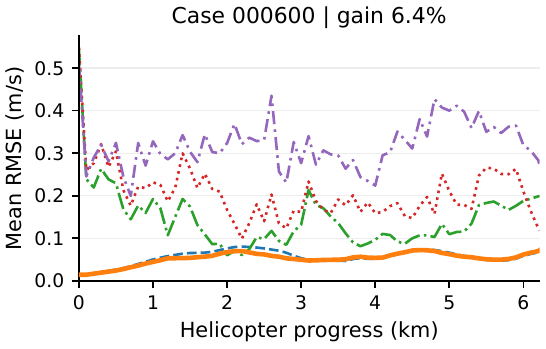}\\[4pt]
\includegraphics[width=.48\linewidth]{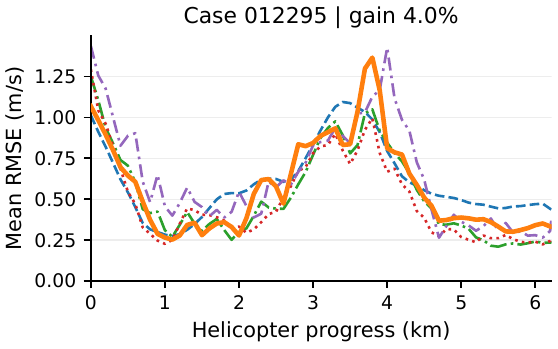}\hfill
\includegraphics[width=.48\linewidth]{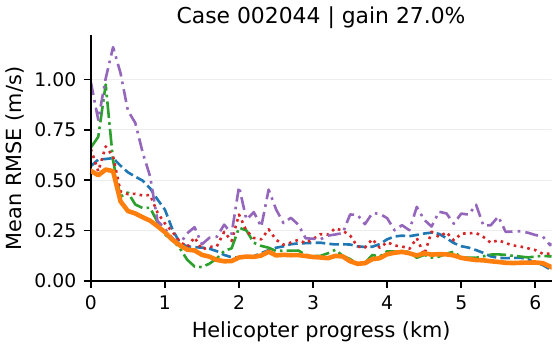}\\[4pt]
\includegraphics[width=.48\linewidth]{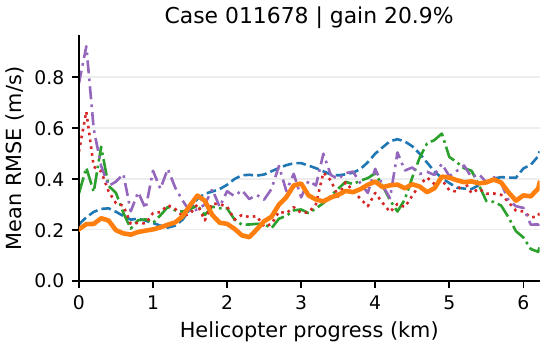}\hfill
\includegraphics[width=.48\linewidth]{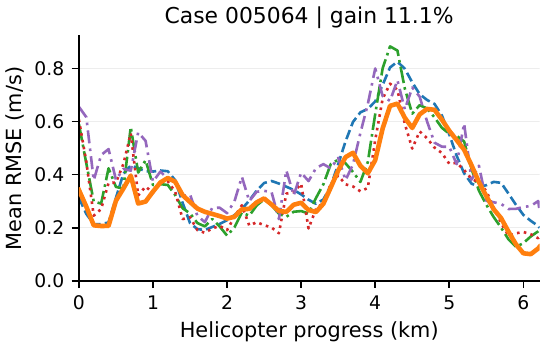}
\caption{\textbf{Complete per-case trajectory diagnostics.} All five methods, all eight cases, and every recorded time step are shown. The primary 50\,m holdout metric is used throughout. Vertical scales differ by case to preserve within-case detail. }
\label{fig:corridor-cases}
\end{figure}

\clearpage
\subsection{Query budget, update cost, and reproducibility limits}
\begin{figure}[ht]
\centering
\includegraphics[width=.49\linewidth]{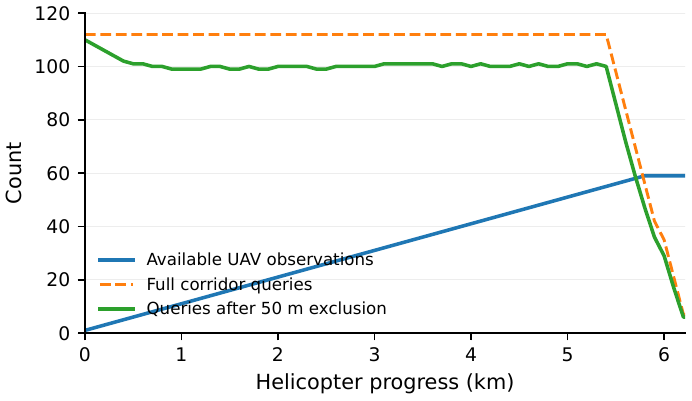}\hfill
\includegraphics[width=.49\linewidth]{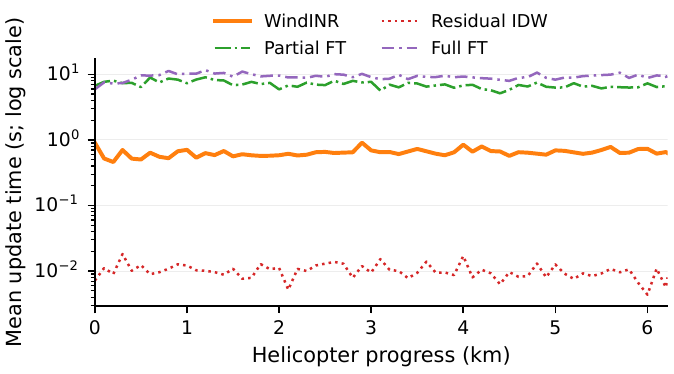}
\caption{\textbf{Available information and measured update cost.} Left: the observation prefix grows to 59 samples while the corridor is clipped near the endpoint; all cases share these counts. Right: mean reported update time across cases at each progress point on a logarithmic axis. }
\label{fig:corridor-budget-cost}
\end{figure}

\begin{table}[htbp]
\centering\small
\caption{Reported update-time distribution across the 512 case--time records (seconds). These values depend on the fixed solver budgets and the code's timing boundaries.}
\label{tab:corridor-timing}
\begin{tabular}{lrrr}
\toprule
Method & Mean & Median & 90th percentile \\
\midrule
WindINR & 0.643 & 0.631 & 0.955 \\
Partial FT & 6.995 & 6.876 & 10.143 \\
Residual IDW & 0.010 & 0.006 & 0.024 \\
Full FT & 9.260 & 8.918 & 13.061 \\

\bottomrule
\end{tabular}
\end{table}

The mean update times are 0.643\,s for WindINR, 6.995\,s for Partial FT, 9.260\,s for Full FT, and 0.010\,s for Residual IDW. Optimization-based timers exclude common corridor decoding; Residual IDW uses a separate interpolation-path timer. Four two-case shards ran on separate NVIDIA A100-PCIE-40GB GPUs, with at most four GPUs active. Reported wall time was 38.5 minutes, summed GPU process time 2.42 GPU-hours, and peak allocated memory 968\,MiB; independent repeated timing trials are unavailable.

The pilot supports mobile observation-driven correction using an already trained representation. Its remaining scope limits are eight cases, one route geometry, a single altitude and lead distance, one noise realization per case, and stationary simulation truth. Correlated route samples, meteorological evolution, different sensing patterns, and out-of-distribution terrains require separate tests.

\clearpage
\subsection{Terrain maps for all eight cases}
\FloatBarrier

\section{Real-UAV observations: protocols and supplementary diagnostics}
\label{app:real-uav}
This appendix supports the three-protocol \tromso{} evaluation in Section~\ref{sec:uav}: one \tromso{} spatial-block test, and two retrospective \tromso{} prefix/suffix tests. 

\subsection{Data sources and preparation}
\label{app:real-uav-data}
\paragraph{\tromso{} episode.}
The evaluated measurements were collected on 4 September 2026 and comprise a horizontal survey and a transect with 15\,s stops. We report one spatial-block test and two prefix/suffix protocols (survey and stopped transect). Timestamps are treated as UTC; acquisition and preparation details are retained in the accompanying manifest.

Available background products are the Nordic 1\,km postprocessed 10\,m analysis and AROME-Arctic fields at 10, 80, and 120\,m. The terrain preparation uses Copernicus GLO-30 resampled to a 10\,m output grid, in EPSG:25833, plus WorldCover-derived roughness. The resampled grid spacing is not the native DEM resolution.

\subsection{Holdouts, metrics, and correction settings}
\label{app:real-uav-protocol}
Each unit uses 16 observation super-observations. The \tromso{} spatial block has 12 holdout super-observations, and the survey and stopped-transect protocols have 19 and 24, respectively. 

The \tromso{} spatial geometry was recorded as locked before its measurements were first scored. Its reported nearest-observation distances have minimum 107.4\,m and median 151.3\,m. The other \tromso{} tests are retrospective, and the stopped-transect suffix consists entirely of revisits within 50\,m of observations.

The primary score is the horizontal-vector RMSE on the held-out measurements,
\begin{equation}
 R_{uv}=\sqrt{\frac{1}{N_h}\sum_{j=1}^{N_h}
 \left[(\widehat u_j-u_j)^2+(\widehat v_j-v_j)^2\right]}
 =\sqrt{R_u^2+R_v^2}.
 \label{eq:real-uav-metric}
\end{equation}
This is not the arithmetic mean of component RMSEs in Eq.~\eqref{eq:metrics}. For the four baselines $\mathcal B=\{\text{Constant},\text{Residual IDW},\text{Partial FT},\text{Full FT}\}$ displayed in Table~\ref{tab:real-uav-all}, define
\begin{equation}
 R_{\rm best}=\min_{m\in\mathcal B}R_m,\qquad
 g=100\,\frac{R_{\rm best}-R_{\rm WindINR}}{R_{\rm best}}.
 \label{eq:real-uav-gain}
\end{equation}
Positive $g$ means lower error than all four baselines. Constant is the strongest comparator on the spatial block and stopped transect, while Partial FT is strongest on the survey; the corresponding gains are 15.84\%, 20.81\%, and 0.67\%, respectively. 

\paragraph{Correction and fine-tuning settings.}
All methods use the same 16 aggregated $u,v$ observations and the same held-out queries. Constant adds one observation-derived horizontal-wind offset to the frozen background at every query. Residual IDW interpolates the observation residuals with inverse-distance exponent 2, horizontal scale 100\,m, and vertical scale 50\,m. Partial FT updates \texttt{decoder.latent\_seed} and the final output-head layer, with parameter-anchor weight 0.01. Full FT updates all network parameters. Both use Adam for 100 steps at learning rate $10^{-4}$, with zero weight decay and no gradient clipping. Only the observation points are used for fitting; held-out measurements are not used for fitting or model selection.

\end{document}